\documentclass[runningheads]{llncs} 
\usepackage{graphicx}
\usepackage{comment}
\usepackage{url}
\usepackage{rotating}

\begin{document}
\title{Harnessing the power of LLMs for normative reasoning in MASs}

\author{Bastin Tony Roy Savarimuthu\inst{1}\orcidID{0000-0003-3213-6319} \and
{Surangika Ranathunga\inst{2}\orcidID{0000-0003-0701-0204}} \and
{Stephen Cranefield\inst{1}\orcidID{0000-0001-5638-1648}}}
\authorrunning{Savarimuthu et al.}
%
\institute{University of Otago, Dunedin, New Zealand\\
\email{(tony.savarimuthu, stephen.cranefield)@otago.ac.nz}\\
 \and
Massey University, Auckland, New Zealand\\
\email{S.Ranathunga@massey.ac.nz}
}%
\maketitle              
\begin{abstract}
Software agents, both human and computational, do not exist in isolation and often need to collaborate or coordinate with others to achieve their goals. In human society, social mechanisms such as norms ensure efficient functioning, and these techniques have been adopted by researchers in multi-agent systems (MAS) to create socially aware agents. However, traditional techniques have limitations, such as operating in limited environments often using brittle symbolic reasoning. The advent of Large Language Models (LLMs) offers a promising solution, providing a rich and expressive vocabulary for norms and enabling \emph{norm-capable} agents that can perform a range of tasks such as norm discovery, normative reasoning and decision-making. This paper examines the potential of LLM-based agents to acquire normative capabilities, drawing on recent Natural Language Processing (NLP) and LLM research. We present our vision for creating \emph{normative LLM agents}. In particular, we discuss how the recently proposed ``LLM agent'' approaches can be extended to implement such normative LLM agents. We also highlight challenges in this emerging field. This paper thus aims to foster collaboration between MAS, NLP and LLM researchers in order to advance the field of normative agents.

\end{abstract}

\section{Introduction \label{sec:intro}} 
In the dynamic field of multi-agent systems (MAS), the integration of Large Language Models (LLMs) with software agents is emerging as a promising area of research~\cite{wang2024survey,xi2023rise}. In traditional MAS, agents are often limited by brittle symbolic reasoning and confined to specific environments. In contrast, agents powered by LLMs (\textit{LLM agents}) have the potential to be equipped with implicit world knowledge. This evolution is crucial for their effective participation in sociotechnical systems, ensuring more efficient coexistence and collaboration between humans and agents in societies. {However, given the novelty of this emerging research area, current LLM-based agent research has not discussed the use of LLMs to assist agents with social knowledge and understanding,
including normative aspects.} This paper explores the potential of equipping LLM-based agents with advanced reasoning skills, particularly in the context of normative decision-making. 

Norms prescribe behaviours that are expected in societies. In normative multi-agent systems, researchers have investigated how human norms can be operationalised in computational societies where an agent's behaviour is constrained by norms \cite{lopez2002}.  However, existing MAS research has primarily focused on agents making normative decisions based on symbolic expressions, which are specific to the formalisms used \cite{choi2022}. 

{Recently, LLM-based reasoning about social norms has gained the attention of Natural Language Processing (NLP) researchers. There is emerging research in using LLMs for norm discovery and moral judgement over textual data such as stories and chat communications~\cite{jiang2021can,ziems2022moral}.} However, so far this strand of research has not been harnessed to implement normative multi-agent systems.

Such NLP research suggests that the advent of LLMs introduces the possibility of employing a rich, expressive vocabulary for norms, enabling agents to gain a range of normative capabilities. Norm discovery, reasoning and conformance are some examples of such normative capabilities. These capabilities are required to seek appropriate knowledge about norms, based on the phase of the norm life-cycle of a particular norm \cite{savarimuthu2010}. In particular, LLMs will enable agents to reason with norms using natural language that includes asking questions such as \emph{what norms apply in a given context?}, \emph{what are the punishments for norm violations?}, and \emph{what action must an agent pursue to be norm compliant?} This development opens avenues for more human-like cognitive abilities in agents, which are essential for collaborative tasks in various domains such as human-robot interaction.

The goal of this paper is to present a vision for creating \textbf{normative LLM agents} that have capabilities such as norm discovery, reasoning and compliance. Our proposal has the potential to instill superior norm capabilities at both the individual agent level and the societal level, overcoming limitations of traditional techniques. This paper also emphasizes the need for further collaboration among MAS, NLP, and LLM researchers to fully realize the potential of normative LLM agents. Such interdisciplinary effort will foster a new era of normatively adept computational agents.

\section{The Promise of LLMs for MAS} \label{section:promise_of_LLMs}
In recent years, the rapidly increasing capabilities of LLMs have astounded both AI practitioners and users, especially since the release of OpenAI's ChatGPT. LLMs are nothing more than complex probabilistic models that calculate the most likely word (or, more accurately, token) given a sequence of previous tokens. However, their massive amount of textual training data and self-supervised training regime means that they exhibit a sophisticated understanding of {syntax and semantics} of natural language. Furthermore, fine-tuning these models with carefully constructed prompts, as well as techniques such as reinforcement learning, can condition the model to produce answers that appear highly relevant to the prompt, giving the impression of human-like understanding.

An emerging and rapidly advancing thread of research on LLMs is addressing the question of how intelligent software agents can leverage the abilities of LLMs to help them make decisions (e.g., choose their actions). A recent survey paper by Wang et al.\@ \cite{wang2024survey} assimilates developments in the literature to propose a standard set of modules required for an LLM-based autonomous agent to effectively leverage an LLM's capabilities: 1) a \emph{profile} module to store information about agent's roles, personalities, relationships, etc., 2) a \emph{memory} module to record the agent's experiences in its environment, 3) a \emph{planning} module to break complex tasks into simpler ones, and 4) an \emph{action} module to translate decisions into actions in the environment. The LLM can then provide the agent with the ability to operate in unconstrained settings unlike the traditional approaches which are usually restricted to specific problem domains. In addition, Xi et al.\@ \cite{xi2023rise} note that ``LLMs can adapt to various languages, cultures, and domains, making them versatile and reducing the need for complex training processes and data collection''.

Furthermore, Andreas et al.\@ \cite{DBLP:conf/emnlp/Andreas22} argues that ``Language Models are models of intentional communication'' in the sense that providing information about agents' beliefs, desires and intention within prompts causes LLMs to generate outputs that appear to relate to these mental attitudes in predictable ways.

As part of a survey of LLM-based agents, Xi et al.\@ \cite{xi2023rise} discuss research on LLM-based agents interacting in a society. Work in this area has included studies on the interactions between agents with individual personalities and emotions, cooperation in teams and spontaneous social interactions. For example, Park et al.\@ \cite{Park2023GenerativeAgents} observe that LLMs ``encode a wide range of human behavior from their training data'' and describe a multi-agent simulation where \emph{generative agents} use LLMs to reflect on their memories to produce high level insights on the activities and motivations of the agents. In combination with memory and planning modules, this approach generates ``believable individual and emergent social behaviors''.

Notably, none of the papers mentioned above discuss the use of LLMs to assist software agents in real time with social knowledge and understanding, including normative aspects. Just as Andreas concluded that LLMs are able to model intentional communication, we posit that LLMs are also capable of revealing rich social understanding of narratives (see Section~\ref{sec:NLP_research_on_norms}). We believe they hold great promise for assisting social agents with normative reasoning, especially in sociotechnical-technical systems where human participants naturally conceive and discuss norms in natural language rather than using symbolic representations.

\section{NLP Research on Social Norms} \label{sec:NLP_research_on_norms}
In the last five years, there has been a surge in interest amongst NLP researchers to study social norms. In particular, a nascent NLP task of norm-related reasoning based on text (chatbot conversations, stories, situation descriptions, etc.) has recently emerged. 
This research can be broadly categorised under norm discovery (or norm generation), normative reasoning and norm conformance. This new-found interest of NLP researchers is not surprising. As AI systems gradually evolve to assist and collaborate with humans, they are expected to be aware of social norms that govern human cultural and social spaces. As described below, with the power of recent advancements in LLMs, there have been promising advances in norm-related reasoning over textual data.

NLP researchers have treated \textit{norm discovery} as the task of identifying applicable norms given the textual description of a situation, and addressed it as a text generation problem. Early approaches relied on human annotated data to fine-tune LLMs in order to train them to generate new norms from unseen situations. Here, the situation has been referred by a single sentence~\cite{forbes2020social} or a conversational setup~\cite{ziems2022moral}. Kiehne et al.\@ \cite{kiehne2022contextualizing} argue that  LLMs fine-tuned with human-annotated norms may have biases.
As a solution, they investigate how to remove or introduce contrasting norms during model training. 
Fung et al.~\cite{fung2022normsage} is the first to omit the need for human annotated data altogether. Instead, they  prompt the LLM for norm discovery over dialogue situations. Li et al.\@ \cite{li2023normdial} bootstrap from an existing set of social norms and prompt ChatGPT to generate new norms. 

\textit{Normative reasoning} has been referred as predicting `moral judgement'~\cite{ziems2023normbank,jiang2021can}. The common approach to training a moral judgement prediction system is to fine-tune an LLM with $situation\_description$, $moral\_judgement$ pairs. In its simplest form, normative reasoning has been treated as a text classification problem, where a textual description of an action gets classified as normative or not~\cite{nahian2020learning}. 
More recent work has explored the possibility of more fine-grained reasoning. For example, Ziems et al.~\cite{ziems2023normbank} categorize a given situation as \textit{ok, expected} or \textit{unexpected}. The publicly available Delphi system\footnote{\url{https://delphi.allenai.org/}}
\cite{jiang2021can} can be prompted with a situation (expressed in a single sentence), and it responds with one of the following judgements: \textit{rude, ok, bad, wrong, expected, shouldn't}. 
However, in Delphi, normative reasoning is done in isolation with no reference to surrounding context. ClarifyDelphi\cite{pyatkin2023clarifydelphi} is a system capable of learning to ask questions to elicit salient context, which overcomes this limitation.

\textit{Norm conformance} (or `value alignment' as termed by NLP researchers) refers to creating agents whose behaviours conform to expected moral and social norms for a given context and a group of people~\cite{ammanabrolu2022aligning}. Emelin et al.~\cite{emelin2021moral} present LLM-based classification and generative models to determine the most suitable action given a situation and a norm. Using the Delphi system~\cite{jiang2021can}, Ammanabrolu et al.\@ \cite{ammanabrolu2022aligning} make an agent learn to constrain its action space such that its actions align with socially beneficial human values. 

In order to train LLMs for the above tasks,  many human-annotated text datasets have been created. Some such datasets are \textsc{Social-Chem-101}~\cite{forbes2020social}, Moral Stories~\cite{emelin2021moral}, \textsc{Scruples}~\cite{lourie2021scruples}, the \textsc{Moral Integrity Corpus}~\cite{ziems2022moral}, \textsc{NormBank}~\cite{ziems2023normbank} and \textsc{NormDial}~\cite{li2023normdial}.

This LLM-based normative reasoning research presents an exciting opportunity to enhance an agent's social reasoning potential. However, none of these work have been harnessed by MAS researchers as a part of an agent's social decision-making engine.  

\section{Implementing Normative LLM agents} \label{sec:normative_LLM_agent}

A normative agent in a traditional MAS is typically created by incorporating domain knowledge expressed in symbolic logical representations \cite{meyer1994,broersen2012}, which is then used for normative reasoning. In many cases, agents using this approach know about the norm a priori. However, such a priori knowledge is unrealistic for agents participating in open systems where norms change. Also, the complexity and nuances associated with norms can make it difficult to model them using formal approaches.

In order to overcome these limitations of traditional normative agents, we propose the MAS community should actively engage in developing normative LLM agents. In particular, we propose to extend the emerging proposals on creating generic LLM agent architectures~\cite{wang2024survey} such that these LLM agents are enriched with norm capabilities.

The key question in developing normative LLM agents is how to train an LLM to reliably demonstrate normative capability (e.g., norm discovery, reasoning and conformance), and integrate them to an agent-based system. We relate these capabilities to NLP-related research on norms discussed in Section \ref{sec:NLP_research_on_norms}. Below, we characterise the nature of these normative capabilities and suggest how these can be realised in the context of building a normative LLM agent. 

\textbf{\emph{Norm discovery - }} First and foremost, the LLM should be equipped with norm-related knowledge. As done by Fung et al.\@ \cite{fung2022normsage} (see Section~\ref{sec:NLP_research_on_norms}), one could prompt a general-purpose LLM to discover norms implicitly captured in its world knowledge. However, generic norms most likely would not work for domain- or culture-specific agent systems. In such cases, the LLM can be further fine-tuned with human-annotated norm data. Note that, as discussed in Section~\ref{sec:NLP_research_on_norms}, this is a common approach to teaching the notion of norms to LLMs. 

A particular limitation of fine-tuning is that it results in a static set of norms (i.e., these don't change). However, norms may not be static and some tend to emerge from complex interactions in an agent society. In order to identify emerging norms, we propose that the \textit{planning module} of Wang et al.\@ \cite{wang2024survey}'s LLM-agent architecture can be extended with normative reasoning capabilities. For example, an agent should be able to identify sanctions based on what it perceives from the environment (e.g. being told off), and then learn norms based on that. Approaches such as Theory of Mind (ToM) for LLM agents~\cite{li2023theory} can be used for this kind of norm identification. 

Another issue that needs to be addressed is to avoid the need for fine-tuning of an LLM every time a new norm is discovered. We suggest extending the memory architectures (such as vector databases) proposed by Wang et al.\@ \cite{wang2024survey} to store such dynamically discovered norms.

\textbf{\emph{Norm reasoning - }} Norm reasoning involves capabilities such as norm identification\footnote{This is somewhat different from norm discovery discussed in the previous sub-section, which aims to discover a set of all possible norms that may apply in a society, while identification refers to reasoning about specific norms that apply in a given context.}, violation detection, sanction identification, norm spreading to other agents and norm enforcement of non-compliant agents. As mentioned above, approaches such as ToM for LLM agents can be explored for violation detection and sanction identification. Norm identification (i.e., which norm currently applies in the given situation) also involves the \textit{planning module} of the LLM agent. While the normative reasoning research in the NLP field (see Section~\ref{sec:NLP_research_on_norms}) would be a good starting point, we envisage more sophisticated modes of LLM usage by a normative agent for explicitly reasoning about a norm (e.g., \emph{what actions are prohibited?}). Some possibilities for norm identification are as follows.

\begin{itemize}
    \item Prompt the LLM with the situation information, without explicitly stating a norm. For example, a robot working with humans can provide the contextual information to the LLM and ask for the norm that applies for its current situation.
    \item Prompt the LLM with the situation information and applicable norm(s), and ask whether the specified norm is applicable to that situation (in order to check its norm understanding).
    \item Prompt the LLM in an iterative manner in order to obtain further normative information, say by asking a series of questions (e.g., using techniques such as Chain of Thought~\cite{wei2022chain})
\end{itemize} 

\textbf{\emph{Norm conformance - }} Once the agent knows which norm applies in a given situation, it should be capable of selecting a norm-adhering action (i.e., to choose to conform to the norm). We propose to extend the \emph{action module} of the LLM agent architecture~\cite{wang2024survey} in order to achieve this. Note that selecting a norm adhering action is very similar to the task of norm conformance (value alignment) discussed in Section~\ref{sec:NLP_research_on_norms}. Instead of generating a textual output as the action as done in that NLP research, here, we expect the agent architecture to generate an action to be executed in the environment that the agent is situated in. One approach to achieve this is by providing specific a priori guidance to the LLMs (i.e., prompting the LLM to choose from a set of norm-aligned actions) using techniques such as Retrieval Augmented Generation (RAG)~\cite{lewis2020retrieval}. The challenges in achieving this are discussed in Section \ref{sec:application}.

While the discussion above only focused on normative LLM agent capabilities related to prior work in NLP (i.e., norm discovery, reasoning and conformance), there is an opportunity to explore the use of LLMs in achieving richer and nuanced normative capabilities for agents. 

\emph{First}, capabilities such as the ability to spread norms through communication and enforce a norm through punishment can be explored in depth. For example, a norm spreading agent should be able convey the norm in a convincing way (e.g., by providing evidence that the norm is in force, and the penalty for non-adherence), and the evidence can come from the LLM.  More generally, this rich set of capabilities can be provided to an agent by enabling the LLM-agent to recognise the current phase of the norm in the norm life-cycle model (e.g., whether a norm is being created, identified, spread and enforced \cite{savarimuthu2011,morris2019,frantz2018}). 

\emph{Second}, certain assumptions made by prior work can be relaxed. Prior work assumes that a sanction is observed in the form of a signal (a symbol \cite{savarimuthu2013}), and the question of where sanctions come from, their different types, and why agents sanction have remained unanswered. With the advent of LLMs, the task of reasoning about sanctions (e.g., seeking explanation) can be delegated to the LLM (after training with appropriate data). Thus, existing gaps in reasoning can be plugged using LLMs. \emph{Third}, norm-relevant explanation can be offered by an LLM agent - e.g., \emph{why does an agent perform action X and not Y in a given normative context?}. Norm-related explanation generation is an area which has been underexplored, and an LLM agent is well-poised to explain the reasons behind action selection and avoidance, corresponding to obligation and prohibition norms respectively. \emph{Fourth}, there is limited prior work that has investigated the dual (slow and fast) nature of normative reasoning, which is routinely employed by humans~\cite{kahneman2017}. The core idea is that some norms are applied fast, based on how often an agent employs it, such as offering a greeting, while some require norm identification from scratch. With the use of LLMs, agents will have the option to do fast reasoning based on its local memory, and consult an LLM for slow reasoning. This presents the opportunity to experiment with different application modes (i.e., when to cache well-known norms and when to consult the LLM), considering the performance goals of the system.

\section{An application of a Normative LLM agent} \label{sec:application}

\begin{sidewaysfigure}
\centering
  \includegraphics[width=\linewidth]{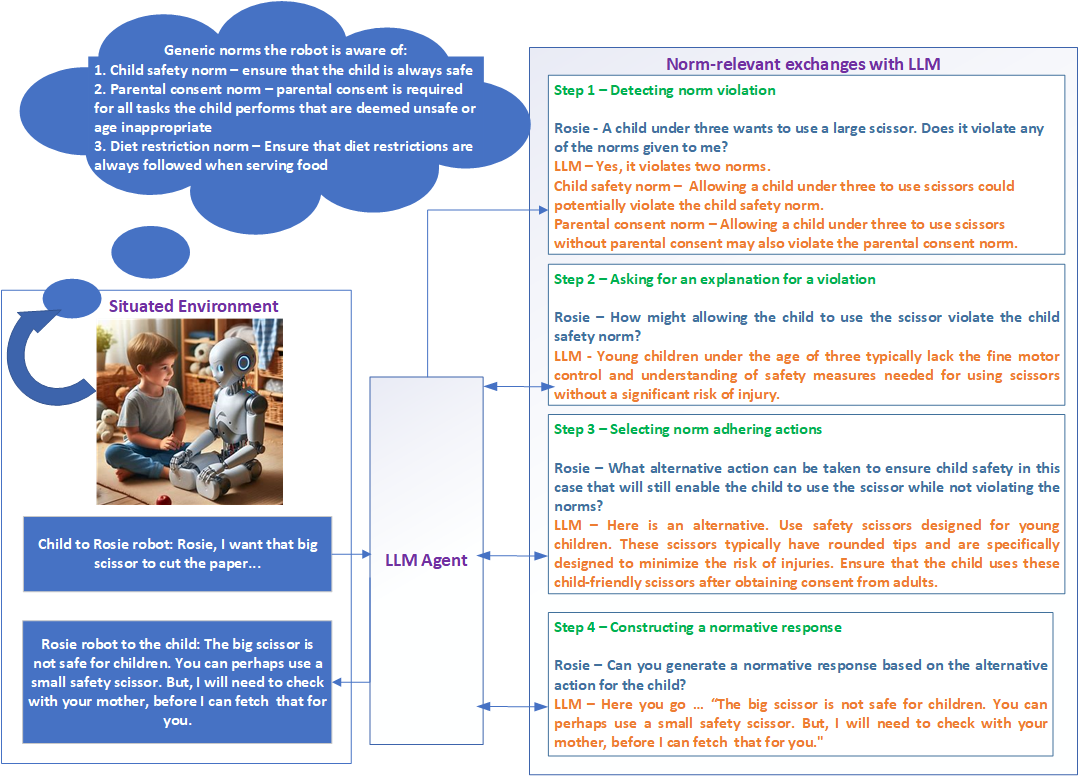}
    \caption{Normative reasoning of a normative LLM agent in a child care scenario}
    \label{fig:scenario}
\end{sidewaysfigure}

We consider a childcare scenario that has been already considered for investigating normative reasoning \cite{canavotto2022,zhang2022}, where Rosie the robot cares for a child in the child's home. Figure \ref{fig:scenario} shows the interactions between the child and Rosie (an embodied normative LLM agent). In this scenario, the child who is less than three years old asks the robot to fetch a large pair of scissors so that he can cut paper and then colour each piece. The robot knows generic norms, as shown inside the thought bubble in the figure, but not all possible instantiations of these generic norms. It consults with the specific LLM model that has been trained on childcare scenarios. The four reasoning steps the agent pursues are shown on the right of Figure \ref{fig:scenario}. The robot consults its LLM module to first check whether there might be norm violations if it were to proceed with the request. Having found that two norms would be violated, it asks for an alternative solution that does not violate the norm, yet satisfies the request from the child. The solution from the LLM is provided by the agent as a response to the child. Once the response is provided, the agent must invoke its planning module to perform the actions that correspond to what has been communicated to the child (i.e., seek consent from the child's mother to use safety scissors and then fetch the scissors and give them to the child). Having learnt the new specific instantiations of the generic norms, it stores them  in its memory for future use. If these specific norms are violated in the future, the agent does need to consult the LLM.

As can be observed, this scenario involves several normative capabilities enabled by the LLM: norm identification through violation detection, seek norm explanation, recommendation of non-violating action and normative response generation. This scenario can be further extended. For example, if the robot was also responsible for entertaining and educating the child, e.g. by playing games, reading books together, then there would be culture specific norms that govern choices of books and games.

Application domains where a normative LLM-agent could prove useful include checking business contract compliance \cite{haque2024},  policy design (e,g., based on the results of simulating norm emergence \cite{ren2024}), and designing virtual organisations where self-regulating agents take preventative or remedial actions when they detect norm conflicts \cite{santos2017}. Other application areas that could be considered include those discussed by the work of Singh et al.\@ \cite{singh2013}. 

\section{Summary and challenges}\label{sec:summary}
This paper proposed to improve normative capabilities of LLMs and to use them in building normative LLM agents, which will enable richer natural language based normative competencies in agent societies. However, we identify five main challenges that need to be addressed to realize their full potential.  

\textbf{1. Retrofitting agent architectures}: Integrating LLMs as a part of the decision-making cycle of an agent may require re-engineering of the existing agent frameworks such as the BDI architecture~\cite{rao1991}. For example, LLMs produce natural language responses that need to be converted back to actions performed by an agent. Thus, in an agent’s planning module, responses need some mapping function that link natural language text to actions (classes, functions etc.). A similar effort is needed to map an agent's perception and beliefs into natural language prompts (e.g.~during normative reasoning). 

There have been some prior work on converting textual observations into logic~\cite{ferraro2020,chaudhury2021,zhan2023} and vice versa \cite{calo2022}, which can be harnessed for this purpose. For example, the work of Chaudhury et al.\@ \cite{chaudhury2021} has proposed a neuro-symbolic approach where an agent learns to perform a specific action (e.g., whether to enter a room or not) based on symbolic abstractions from textual observations (i.e., description of the state of the room). However, these works have not considered the use of LLMs for this purpose. 

\textbf{2. Inherent capability limitations of current LLMs}:
As probabilistic models trained to predict the next token, LLMs are prone to hallucination and may produce inaccurate facts~\cite{huang2023survey}. Many researchers have argued (and empirically shown) that LLMs are still far behind human reasoning capabilities~\cite{yiu2023transmission,mialon2024gaia}. This suggests that a substantial effort needs to be employed to teach human-level normative reasoning capabilities to LLMs. Moreover, capability to obtain proper responses from an LLM requires a carefully generated prompt, which requires human expertise. 

\textbf{3. Resource constraints}: LLMs are prohibitively expensive to train: they need large GPU clusters and training expands over days to months. Thus it is not practical for an average researcher to build custom LLMs from scratch. Even fine-tuning LLMs requires a computer with a GPU that is well-equipped with memory. Prompting proprietary LLMs such as ChatGPT incurs non-trivial amounts of time and money \cite{Park2023GenerativeAgents}. 

\textbf{4. Data constraints}: To fine-tune LLMs for specific agent scenarios, human-annotated domain data and also culture-specific data may be needed. The domain experts needed to curate domain-specific data may be costly. Obtaining culture-specific data is even harder. Also, there is a significant linguistic disparity in the field of NLP~\cite{ranathunga2022some}, and the performance of LLMs for low-resource languages is far behind that of their high-resource counterparts~\cite{lee2022pre}. 

\textbf{5. Ethical issues}: Using LLMs require careful consideration as a range of ethical issues need to be addressed including privacy, transparency, reproducibility, explainability, and potential data biases. In particular, using AI systems for moral reasoning has been criticized by some researchers~\cite{talat2022machine}. Thus, integrating frameworks that enable systematic mitigation of ethical issues is required \cite{watkins2023}. 

It is worth noting that NLP and LLM researchers are advancing in an unprecedented speed to address the LLM-related limitations. For example, solutions such as RAG~\cite{lewis2020retrieval} have been employed to overcome the problem of hallucination, and Low-rank Adaptation (LoRA)~\cite{hu2022lora} has been very successful in reducing the memory requirements during LLM fine-tuning. Thus, we expect that most of the LLM-related challenges to be resolved to a greater extent in the near future. We invite MAS researchers to tackle challenges on the MAS side (e.g., creating integrated architectures) in order to harness the potential of LLMs to create socially-situated, expressive and norm-competent agents. Our proposal here complements the proposal of Wells and Snaith \cite{wells2023} on using LLMs and dialogue based systems to deliberate about norms.

In the short term, we suggest that researchers focus on specific agendas such as a) investigating how an LLM-agent can employ sanctions \cite{villatoro2010} and reputation mechanisms \cite{santos2018} to facilitate norm emergence, and b) exploring how norm entrepreneurship \cite{anavankot2024} can be facilitated by an LLM-agent, and how these norms may go on to become established norms, and then (potentially) superseded by other norms based on changing circumstances. The recent work of Ren et al.\@ \cite{ren2024} demonstrates how many phases of the norm-life cycle model can be simulated using LLM-based normative agents, and their framework can be extended to explore the emergence of complex and conditional norms. These studies can demonstrate the potential for the LLM-agent to demonstrate real-life norm emergence processes.

Beyond creating normative LLM-agents which has been the focus of this paper (and there have been efforts in this direction \cite{he2024,ren2024,haque2024}), LLMs can also be useful in enabling smoother functioning of a normative system where it can play the role of an overseer or a governor agent that monitors normative activities of an agent society. For example,  LLMs can be useful for detecting the specific life-cycle phase of a norm (whether a norm is in the emergent phase, established phase or declining phase \cite{savarimuthu2013b}) based on observing all interactions in an agent society. Additionally, LLMs can be used to synthesise and propose new norms that are more efficient that current norms \cite{morales2014}, detect norm conflicts at the societal level and offer conflict resolution mechanisms \cite{vasconcelos2009}. We anticipate such developments will be beneficial for the creation and embedding of norm-capable agents (e.g., robots) in real world settings.




\bibliographystyle{splncs04}
\bibliography{coine_blue_sky}

\begin{thebibliography}{10}
\providecommand{\url}[1]{\texttt{#1}}
\providecommand{\urlprefix}{URL }
\providecommand{\doi}[1]{https://doi.org/#1}

\bibitem{ammanabrolu2022aligning}
Ammanabrolu, P., Jiang, L., Sap, M., Hajishirzi, H., Choi, Y.: Aligning to social norms and values in interactive narratives. In: Proceedings of the 2022 Conference of the North American Chapter of the Association for Computational Linguistics: Human Language Technologies. pp. 5994--6017 (2022)

\bibitem{anavankot2024}
Anavankot, A.M., Cranefield, S., Savarimuthu, B.T.R.: {NEMAS}: Norm entrepreneurship in multi-agent systems. Systems  \textbf{12}(6), ~187 (2024)

\bibitem{DBLP:conf/emnlp/Andreas22}
Andreas, J.: Language models as agent models. In: Goldberg, Y., Kozareva, Z., Zhang, Y. (eds.) Findings of the Association for Computational Linguistics: {EMNLP} 2022, Abu Dhabi, United Arab Emirates, December 7-11, 2022. pp. 5769--5779. Association for Computational Linguistics (2022). \doi{10.18653/v1/2022.findings-emnlp.423}

\bibitem{broersen2012}
Broersen, J., van~der Torre, L.: Ten problems of deontic logic and normative reasoning in computer science. Lectures on Logic and Computation: ESSLLI 2010 Copenhagen, Denmark, August 2010, ESSLLI 2011, Ljubljana, Slovenia, August 2011, Selected Lecture Notes pp. 55--88 (2012)

\bibitem{calo2022}
Cal{\`o}, E., van~der Werf, E., Gatt, A., van Deemter, K.: Enhancing and evaluating the grammatical framework approach to logic-to-text generation. In: Proceedings of the 2nd Workshop on Natural Language Generation, Evaluation, and Metrics (GEM). pp. 148--171 (2022)

\bibitem{canavotto2022}
Canavotto, I., Horty, J.: Piecemeal knowledge acquisition for computational normative reasoning. In: Proceedings of the 2022 AAAI/ACM Conference on AI, Ethics, and Society. pp. 171--180 (2022)

\bibitem{chaudhury2021}
Chaudhury, S., Sen, P., Ono, M., Kimura, D., Tatsubori, M., Munawar, A.: Neuro-symbolic approaches for text-based policy learning. In: Proceedings of the 2021 Conference on Empirical Methods in Natural Language Processing. pp. 3073--3078 (2021)

\bibitem{choi2022}
Choi, Y.: The curious case of commonsense intelligence. Daedalus  \textbf{151}(2),  139--155 (2022)

\bibitem{emelin2021moral}
Emelin, D., Le~Bras, R., Hwang, J.D., Forbes, M., Choi, Y.: {Moral Stories}: Situated reasoning about norms, intents, actions, and their consequences. In: Proceedings of the 2021 Conference on Empirical Methods in Natural Language Processing. pp. 698--718 (2021)

\bibitem{ferraro2020}
Ferraro, G., Lam, H.P., Tosatto, S.C., Olivieri, F., Islam, M.B., van Beest, N., Governatori, G.: Automatic extraction of legal norms: Evaluation of natural language processing tools. In: New Frontiers in Artificial Intelligence: JSAI-isAI International Workshops, JURISIN, AI-Biz, LENLS, Kansei-AI, Yokohama, Japan, November 10--12, 2019, Revised Selected Papers 10. pp. 64--81. Springer (2020)

\bibitem{forbes2020social}
Forbes, M., Hwang, J.D., Shwartz, V., Sap, M., Choi, Y.: {\scshape Social Chemistry 101}: Learning to reason about social and moral norms. In: Proceedings of the 2020 Conference on Empirical Methods in Natural Language Processing (EMNLP). pp. 653--670 (2020)

\bibitem{frantz2018}
Frantz, C., Pigozzi, G.: Modeling norm dynamics in multiagent systems. Journal of Applied Logic  \textbf{5}(2),  491--564 (2018)

\bibitem{fung2022normsage}
Fung, Y., Chakrabarty, T., Guo, H., Rambow, O., Muresan, S., Ji, H.: {\scshape NormSage}: Multi-lingual multi-cultural norm discovery from conversations on-the-fly. In: Bouamor, H., Pino, J., Bali, K. (eds.) Proceedings of the 2023 Conference on Empirical Methods in Natural Language Processing. pp. 15217--15230. Association for Computational Linguistics, Singapore (Dec 2023). \doi{10.18653/v1/2023.emnlp-main.941}

\bibitem{haque2024}
Haque, A., Singh, M.P.: Extracting norms from contracts via chatgpt: Opportunities and challenges. arXiv preprint arXiv:2404.02269  (2024)

\bibitem{he2024}
He, S., Ranathunga, S., Cranefield, S., Savarimuthu, B.T.R.: Norm violation detection in multi-agent systems using large language models: A pilot study. arXiv preprint arXiv:2403.16517  (2024)

\bibitem{hu2022lora}
Hu, E.J., Shen, Y., Wallis, P., Allen-Zhu, Z., Li, Y., Wang, S., Wang, L., Chen, W.: Lo{RA}: Low-rank adaptation of large language models. In: International Conference on Learning Representations (2022), \url{https://openreview.net/forum?id=nZeVKeeFYf9}

\bibitem{huang2023survey}
Huang, L., Yu, W., Ma, W., Zhong, W., Feng, Z., Wang, H., Chen, Q., Peng, W., Feng, X., Qin, B., et~al.: A survey on hallucination in large language models: Principles, taxonomy, challenges, and open questions. arXiv preprint arXiv:2311.05232  (2023)

\bibitem{jiang2021can}
Jiang, L., Hwang, J.D., Bhagavatula, C., Bras, R.L., Liang, J., Dodge, J., Sakaguchi, K., Forbes, M., Borchardt, J., Gabriel, S., et~al.: Can machines learn morality? the {Delphi} experiment. arXiv preprint arXiv:2110.07574  (2021)

\bibitem{kahneman2017}
Kahneman, D.: Thinking, fast and slow. Farrar, Straus and Giroux (2017)

\bibitem{kiehne2022contextualizing}
Kiehne, N., Kroll, H., Balke, W.T.: Contextualizing language models for norms diverging from social majority. In: Findings of the Association for Computational Linguistics: EMNLP 2022. pp. 4620--4633 (2022)

\bibitem{lee2022pre}
Lee, E.S., Thillainathan, S., Nayak, S., Ranathunga, S., Adelani, D., Su, R., Mccarthy, A.D.: Pre-trained multilingual sequence-to-sequence models: A hope for low-resource language translation? In: Findings of the Association for Computational Linguistics: ACL 2022. pp. 58--67 (2022)

\bibitem{lewis2020retrieval}
Lewis, P., Perez, E., Piktus, A., Petroni, F., Karpukhin, V., Goyal, N., K{\"u}ttler, H., Lewis, M., Yih, W.t., Rockt{\"a}schel, T., et~al.: Retrieval-augmented generation for knowledge-intensive {NLP} tasks. Advances in Neural Information Processing Systems  \textbf{33},  9459--9474 (2020)

\bibitem{li2023theory}
Li, H., Chong, Y., Stepputtis, S., Campbell, J.P., Hughes, D., Lewis, C., Sycara, K.: Theory of mind for multi-agent collaboration via large language models. In: Proceedings of the 2023 Conference on Empirical Methods in Natural Language Processing. pp. 180--192 (2023)

\bibitem{li2023normdial}
Li, O., Subramanian, M., Saakyan, A., CH-Wang, S., Muresan, S.: {\scshape NormDial}: A comparable bilingual synthetic dialog dataset for modeling social norm adherence and violation. In: Bouamor, H., Pino, J., Bali, K. (eds.) Proceedings of the Conference on Empirical Methods in Natural Language Processing. pp. 15732--15744. Association for Computational Linguistics, Singapore (Dec 2023). \doi{10.18653/v1/2023.emnlp-main.974}

\bibitem{lopez2002}
L{\'o}pez~y L{\'o}pez, F., Luck, M., d'Inverno, M.: Constraining autonomy through norms. In: Proceedings of the First International Joint Conference on Autonomous Agents and Multiagent Systems: Part 2. pp. 674--681 (2002)

\bibitem{lourie2021scruples}
Lourie, N., Le~Bras, R., Choi, Y.: {\scshape Scruples}: A corpus of community ethical judgments on 32,000 real-life anecdotes. In: Proceedings of the AAAI Conference on Artificial Intelligence. vol.~35, pp. 13470--13479 (2021)

\bibitem{meyer1994}
Meyer, J.J.C., Wieringa, R.J.: Deontic logic in computer science: normative system specification. John Wiley \& Sons, Inc. (1994)

\bibitem{mialon2024gaia}
Mialon, G., Fourrier, C., Wolf, T., LeCun, Y., Scialom, T.: {GAIA}: a benchmark for general {AI} assistants. In: The Twelfth International Conference on Learning Representations (2024), \url{https://openreview.net/forum?id=fibxvahvs3}

\bibitem{morales2014}
Morales, J., L{\'{o}}pez{-}S{\'{a}}nchez, M., Rodr{\'{\i}}guez{-}Aguilar, J.A., Wooldridge, M.J., Vasconcelos, W.W.: Minimality and simplicity in the on-line automated synthesis of normative systems. In: Bazzan, A.L.C., Huhns, M.N., Lomuscio, A., Scerri, P. (eds.) International conference on Autonomous Agents and Multi-Agent Systems, {AAMAS} '14, Paris, France, May 5-9, 2014. pp. 109--116. {IFAAMAS/ACM} (2014), \url{http://dl.acm.org/citation.cfm?id=2615752}

\bibitem{morris2019}
Morris-Martin, A., De~Vos, M., Padget, J.: Norm emergence in multiagent systems: a viewpoint paper. Autonomous Agents and Multi-Agent Systems  \textbf{33},  706--749 (2019)

\bibitem{nahian2020learning}
Nahian, M.S.A., Frazier, S., Riedl, M., Harrison, B.: Learning norms from stories: A prior for value aligned agents. In: Proceedings of the AAAI/ACM Conference on AI, Ethics, and Society. pp. 124--130 (2020)

\bibitem{Park2023GenerativeAgents}
Park, J.S., O'Brien, J.C., Cai, C.J., Morris, M.R., Liang, P., Bernstein, M.S.: Generative agents: Interactive simulacra of human behavior. In: 36th Annual ACM Symposium on User Interface Software and Technology (UIST '23). ACM (2023)

\bibitem{pyatkin2023clarifydelphi}
Pyatkin, V., Hwang, J.D., Srikumar, V., Lu, X., Jiang, L., Choi, Y., Bhagavatula, C.: {ClarifyDelphi}: Reinforced clarification questions with defeasibility rewards for social and moral situations. In: Proceedings of the 61st Annual Meeting of the Association for Computational Linguistics (Volume 1: Long Papers). pp. 11253--11271 (2023)

\bibitem{ranathunga2022some}
Ranathunga, S., de~Silva, N.: Some languages are more equal than others: Probing deeper into the linguistic disparity in the {NLP} world. In: Proceedings of the 2nd Conference of the Asia-Pacific Chapter of the Association for Computational Linguistics and the 12th International Joint Conference on Natural Language Processing. pp. 823--848 (2022)

\bibitem{rao1991}
Rao, A.S., Georgeff, M.P.: Modeling rational agents within a {BDI}-architecture. In: Proceedings of the Second International Conference on Principles of Knowledge Representation and Reasoning. p. 473–484. Morgan Kaufmann Publishers Inc. (1991)

\bibitem{ren2024}
Ren, S., Cui, Z., Song, R., Wang, Z., Hu, S.: Emergence of social norms in large language model-based agent societies. arXiv preprint arXiv:2403.08251  (2024)

\bibitem{santos2018}
Santos, F.P., Santos, F.C., Pacheco, J.M.: Social norm complexity and past reputations in the evolution of cooperation. Nature  \textbf{555}(7695),  242--245 (2018)

\bibitem{santos2017}
Santos, J.S., Zahn, J.O., Silvestre, E.A., Silva, V.T., Vasconcelos, W.W.: Detection and resolution of normative conflicts in multi-agent systems: a literature survey. Autonomous agents and multi-agent systems  \textbf{31},  1236--1282 (2017)

\bibitem{savarimuthu2011}
Savarimuthu, B.T.R., Cranefield, S.: Norm creation, spreading and emergence: A survey of simulation models of norms in multi-agent systems. Multiagent and Grid Systems  \textbf{7}(1),  21--54 (2011)

\bibitem{savarimuthu2010}
Savarimuthu, B.T.R., Cranefield, S., Purvis, M.A., Purvis, M.K.: Obligation norm identification in agent societies. Journal of Artificial Societies and Social Simulation  \textbf{13}(4), ~3 (2010)

\bibitem{savarimuthu2013}
Savarimuthu, B.T.R., Cranefield, S., Purvis, M.A., Purvis, M.K.: Identifying prohibition norms in agent societies. Artificial Intelligence and Law  \textbf{21},  1--46 (2013)

\bibitem{savarimuthu2013b}
Savarimuthu, B.T.R., Padget, J., Purvis, M.A.: Social norm recommendation for virtual agent societies. In: PRIMA 2013: Principles and Practice of Multi-Agent Systems: 16th International Conference, Dunedin, New Zealand, December 1-6, 2013. Proceedings 16. pp. 308--323. Springer (2013)

\bibitem{singh2013}
Singh, M.P., Arrott, M., Balke, T., Chopra, A.K., Christiaanse, R., Cranefield, S., Dignum, F., Eynard, D., Farcas, E., Fornara, N., Gandon, F., Governatori, G., Khanh~Dam, H., Hulstijn, J., Krueger, I., Lam, H.P., Meisinger, M., Noriega, P., Savarimuthu, B.T.R., Tadanki, K., Verhagen, H., Villata, S.: {The Uses of Norms}. In: Andrighetto, G., Governatori, G., Noriega, P., van~der Torre, L.W.N. (eds.) Normative Multi-Agent Systems, Dagstuhl Follow-Ups, vol.~4, pp. 191--229. Schloss Dagstuhl -- Leibniz-Zentrum f{\"u}r Informatik, Dagstuhl, Germany (2013)

\bibitem{talat2022machine}
Talat, Z., Blix, H., Valvoda, J., Ganesh, M.I., Cotterell, R., Williams, A.: On the machine learning of ethical judgments from natural language. In: Proceedings of the 2022 Conference of the North American Chapter of the Association for Computational Linguistics: Human Language Technologies. pp. 769--779 (2022)

\bibitem{vasconcelos2009}
Vasconcelos, W.W., Kollingbaum, M.J., Norman, T.J.: Normative conflict resolution in multi-agent systems. Autonomous agents and multi-agent systems  \textbf{19},  124--152 (2009)

\bibitem{villatoro2010}
Villatoro, D., Sen, S., Sabater-Mir, J.: Of social norms and sanctioning: A game theoretical overview. International Journal of Agent Technologies and Systems (IJATS)  \textbf{2}(1),  1--15 (2010)

\bibitem{wang2024survey}
Wang, L., Ma, C., Feng, X., Zhang, Z., Yang, H., Zhang, J., Chen, Z., Tang, J., Chen, X., Lin, Y., et~al.: A survey on large language model based autonomous agents. Frontiers of Computer Science  \textbf{18}(6),  1--26 (2024)

\bibitem{watkins2023}
Watkins, R.: Guidance for researchers and peer-reviewers on the ethical use of large language models ({LLMs}) in scientific research workflows. AI and Ethics pp.~1--6 (2023)

\bibitem{wei2022chain}
Wei, J., Wang, X., Schuurmans, D., Bosma, M., Xia, F., Chi, E., Le, Q.V., Zhou, D., et~al.: Chain-of-thought prompting elicits reasoning in large language models. Advances in Neural Information Processing Systems  \textbf{35},  24824--24837 (2022)

\bibitem{wells2023}
Wells, S., Snaith, M.: On the role of dialogue models in the age of large language models - extended abstract. In: Grasso, F., Green, N.L., Schneider, J., Wells, S. (eds.) Proceedings of the 23nd Workshop on Computational Models of Natural Argument {(CMNA} 2023), Virtual Event, December 3rd, 2023. {CEUR} Workshop Proceedings, vol.~3614, pp. 49--51. CEUR-WS.org (2023)

\bibitem{xi2023rise}
Xi, Z., Chen, W., Guo, X., He, W., Ding, Y., Hong, B., Zhang, M., Wang, J., Jin, S., Zhou, E., Zheng, R., Fan, X., Wang, X., Xiong, L., Zhou, Y., Wang, W., Jiang, C., Zou, Y., Liu, X., Yin, Z., Dou, S., Weng, R., Cheng, W., Zhang, Q., Qin, W., Zheng, Y., Qiu, X., Huang, X., Gui, T.: The rise and potential of large language model based agents: A survey. arXiv preprint arXiv:2309.07864  (2023)

\bibitem{yiu2023transmission}
Yiu, E., Kosoy, E., Gopnik, A.: Transmission versus truth, imitation versus innovation: What children can do that large language and language-and-vision models cannot (yet). Perspectives on Psychological Science p. 17456916231201401 (2023)

\bibitem{zhan2023}
Zhan, N., Sarkadi, S., Such, J.: Privacy-enhanced personal assistants based on dialogues and case similarity. In: European Conference on Artificial Intelligence. IOS Press (2023)

\bibitem{zhang2022}
Zhang, Z.: Research on child care robot and the influence on children. In: 2022 5th International Conference on Humanities Education and Social Sciences (ICHESS 2022). pp. 225--233. Atlantis Press (2022)

\bibitem{ziems2023normbank}
Ziems, C., Dwivedi-Yu, J., Wang, Y.C., Halevy, A., Yang, D.: {\scshape NormBank}: A knowledge bank of situational social norms. In: Rogers, A., Boyd-Graber, J., Okazaki, N. (eds.) Proceedings of the 61st Annual Meeting of the Association for Computational Linguistics (Volume 1: Long Papers). pp. 7756--7776. Association for Computational Linguistics, Toronto, Canada (Jul 2023). \doi{10.18653/v1/2023.acl-long.429}

\bibitem{ziems2022moral}
Ziems, C., Yu, J., Wang, Y.C., Halevy, A., Yang, D.: The {\scshape moral integrity corpus}: A benchmark for ethical dialogue systems. In: Proceedings of the 60th Annual Meeting of the Association for Computational Linguistics (Volume 1: Long Papers). pp. 3755--3773 (2022)

\end{thebibliography}

\end{document}